\documentclass[runningheads]{llncs}
\usepackage[T1]{fontenc}
\usepackage{booktabs}
\usepackage{graphicx}
\usepackage{caption}
\usepackage{subcaption}
\usepackage[normalem]{ulem}
\usepackage{gensymb}
\usepackage{amssymb}
\usepackage{pifont}
\usepackage{multirow}
\newcommand{\cmark}{\ding{51}}
\newcommand{\xmark}{\ding{55}}

\usepackage{todonotes}
\begin{document}
\title{Vision-based Xylem Wetness Classification in Stem Water Potential Determination}
%
%
\author{Pamodya Peiris \inst{1}\orcidID{0000-0003-0318-5971} \and
Aritra Samanta \inst{1}\orcidID{0009-0006-3915-0649} \and
Caio Mucchiani \inst{1}\orcidID{0000-0001-5471-9270} \and
Cody Simons \inst{1}\orcidID{0009-0009-3363-7538}  \and
Amit Roy-Chowdhury \inst{1}\orcidID{0000-0001-6690-9725} \and
Konstantinos Karydis \inst{1}\orcidID{0000-0002-1144-8260}
}

\authorrunning{P. Peiris et al.}
\titlerunning{Vision-based Xylem Wetness Classification}
\institute{University of California, Riverside}
%
\institute{University of California, Riverside, Riverside CA 92521, USA 
\email{\{ppeir002,asama004,caiocesr,csimo005,amitrc,karydis\}@ucr.edu}}

\maketitle              
\begin{abstract}
Water is often overused in irrigation, making efficient management of it crucial. Precision Agriculture emphasizes tools like stem water potential (SWP) analysis for better plant status determination. However, such tools often require labor-intensive in-situ sampling. Automation and machine learning can streamline this process and enhance outcomes. This work focused on automating stem detection and xylem wetness classification using the Scholander Pressure Chamber, a widely used but demanding method for SWP measurement. The aim was to refine stem detection and develop computer-vision-based methods to better classify water emergence at the xylem. To this end, we collected and manually annotated video data, applying vision- and learning-based methods for detection and classification. Additionally, we explored data augmentation and fine-tuned parameters to identify the most effective models. The identified best-performing models for stem detection and xylem wetness classification were evaluated end-to-end over 20 SWP measurements. 
Learning-based stem detection via YOLOv8n combined with ResNet50-based classification achieved a Top-1 accuracy of 80.98\%, making it the best-performing approach for xylem wetness classification.

\keywords{Precision Agriculture \and Automation \and Stem Water Potential \and Stem Detection \and Wetness Classification}
\end{abstract}
%
%
%

\section{Introduction}
Different crops exhibit different characteristics based on the provided amount of water~\cite{shackel2021establishing}. 
While for certain types of crops, introducing a deficit can have potential advantages, for others it may have detrimental effects. 
For example, in one study it was shown untimely irrigation scheduling of avocado plants may lead to plant stress and eventually decreased yields~\cite{michelakis1993water}. 
Another study demonstrated that controlled deficit irrigation in oranges led to increased productivity of about 4\%~\cite{zapata2017controlled}. 
Refining irrigation practices is thus key to sustainable water usage~\cite{oster2003economic}. 
Several tools are available for assessing plant water status, such as measuring soil salinity~\cite{muhammad2023soil} and sap flow~\cite{nadezhdina1999sap}, and proximal sensing of stem water potential~\cite{mucchiani2024assessing}.

Stem water potential (SWP) is widely adopted as an indicator of water stress in plants~\cite{mccutchan1992stem}, reflecting the tension of water within the plant’s stem and serving as a sensitive measure to guide irrigation practices. 
It is considered one of the most accurate plant-based water status measures, especially for fruit trees and vines~\cite{levin2019re}. 
SWP is directly related to water availability and plant transpiration, making it a crucial parameter for agricultural applications, including precision irrigation. 
The pressure chamber is a widely used tool for measuring SWP in plants~\cite{boyer1967leaf}. 
To determine SWP using the pressure chamber, a leaf is first enclosed inside a retroflective bag and left to settle for about 10 minutes. 
The leaf is then cut at its stem from the plant and placed inside the chamber with the cut end protruding. 
Pressurized gas, usually nitrogen, is gradually introduced until water just begins to appear at the cut surface.
This pressure counteracts the suction force within the xylem, which pulls water up from the roots due to transpiration. 
The applied pressure at which water appears is then used as a proxy of the plant’s water potential at the time of sampling. 
This method is particularly effective for assessing plant water stress, as higher pressure values correspond to greater water tension and suggest a water scarcity in the plant~\cite{Suter2019}.

The pressure chamber method shows significant potential for automation~\cite{mucchiani2024assessing}, with one core component being the integration of machine vision and learning to classify xylem wetness in an automated manner. 
A neural network was able to classify the stem as either ``dry'' or ``wet'' in manual~\cite{amel2023} and automated~\cite{mucchiani2024development} pressure chambers. 
Those works considered only two classes, while leaf samples were taken at relatively similar agronomic conditions. 
However, the efficacy of the network considered in these works may be challenged when leaf samples are taken at varied agronomic conditions (most crucially, at distinct arid/irrigated field cycles~\cite{mucchiani2024development}). 
It also does not capture the presence of bubbles that often occur in the SWP process~\cite{wei2000transmission}, which, as we show in this work, can significantly challenge visual learning methods. 
Together, these challenges raise the key question we aim to address in this work: \emph{How to enable generalizable machine vision and learning tools for automated stem detection and xylem wetness classification?} 

Conceptually-related works have studied leaf and disease classification~\cite{al2010framework}, xylem tissue segmentation~\cite{garcia2018xylem,garcia2020convolutional}, and xylem cell type classification~\cite{wu2022deep} using convolutional neural networks (CNNs). 
Classical methods have been used for plant stress detection (e.g., segmentation and gradient-based decision trees~\cite{zhuang2017early} or SWP measurements via leaf length~\cite{yamane2023stem}). 
Given the limited research on detection and classification methods for water observations specifically at the stem cross-section, this study conducts a comprehensive comparative analysis of vision- and learning-based methodologies to identify the distinct observable stages of the stem under increasing pressure inside the chamber. 
In addition to the ``wet'' and ``dry'' classes considered before~\cite{amel2023,mucchiani2024development}, we also consider a third class ``bubble.'' 

\begin{figure}[!t]
    \centering
    \includegraphics[width=0.7\textwidth]{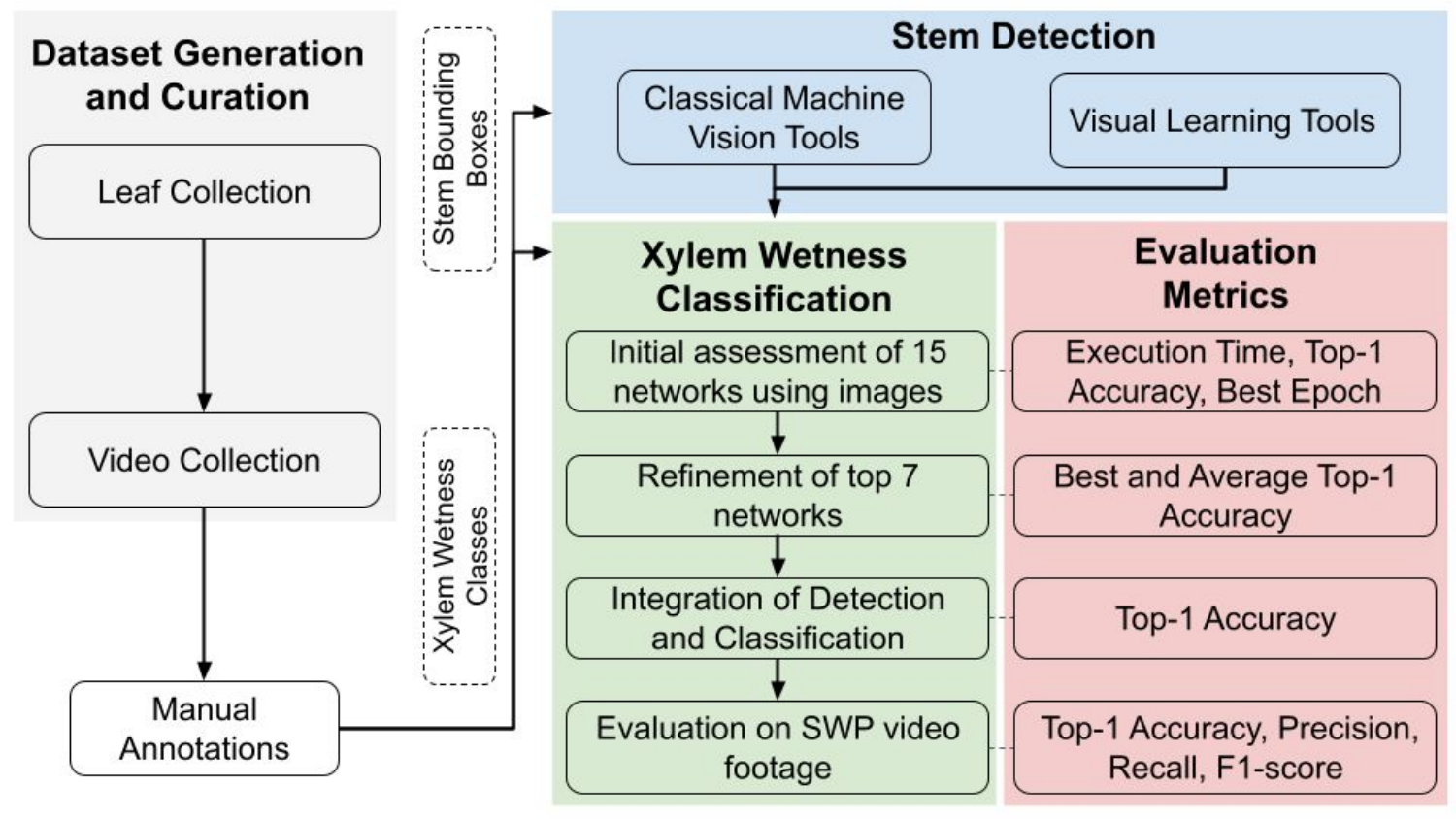}
    \vspace{-6pt}
    \caption{Our end-to-end framework for vision-based xylem wetness classification has three distinct phases. Data collection, and dataset generation and curation help train and assess methods for stem detection. The best-performing stem detection methods are then used together with the dataset to train and evaluate several plausible learning-based xylem wetness classification methods.}
    \label{fig:framework}
    \vspace{-6pt}
\end{figure}

Our framework (Fig.~\ref{fig:framework}) is structured into three phases: data collection, stem detection, and image/video classification. 
We compare stem detection efficacy of classical machine vision (Hough Transforms) and visual learning tools (`You Only Look Once' -- YOLO~\cite{redmon2016you}) in terms of the intersection-over-union (IoU) against manually annotated images, as well as execution time. 
We then evaluate state-of-the-art CNN models and identify the best methods based on accuracy and classification rates for xylem wetness. 
Best-performing detection and classification methods are downselected and tested on SWP measurement video footage.  
Overall, we evaluate fifteen CNN architectures and five detection methods to find that learning-based stem cropping paired with ResNet-50 provides the best Top-1 accuracy in classifying the xylem wetness. 
To our knowledge, this study is the first to comparatively evaluate various detection and classification methods for stem xylem wetness in SWP. 
This step is essential for enhancing precision and accuracy and advancing automated solutions for the pressure chamber method. 

\section{Materials and Methods}

This section outlines all components of this study. 
It describes the experimental setup and data collection procedures, dataset generation and curation, and the methods considered for stem detection and xylem wetness classification.

\subsection{Experimental Setup}

\begin{figure}[!t]
    \centering
    \includegraphics[width=0.90\textwidth]{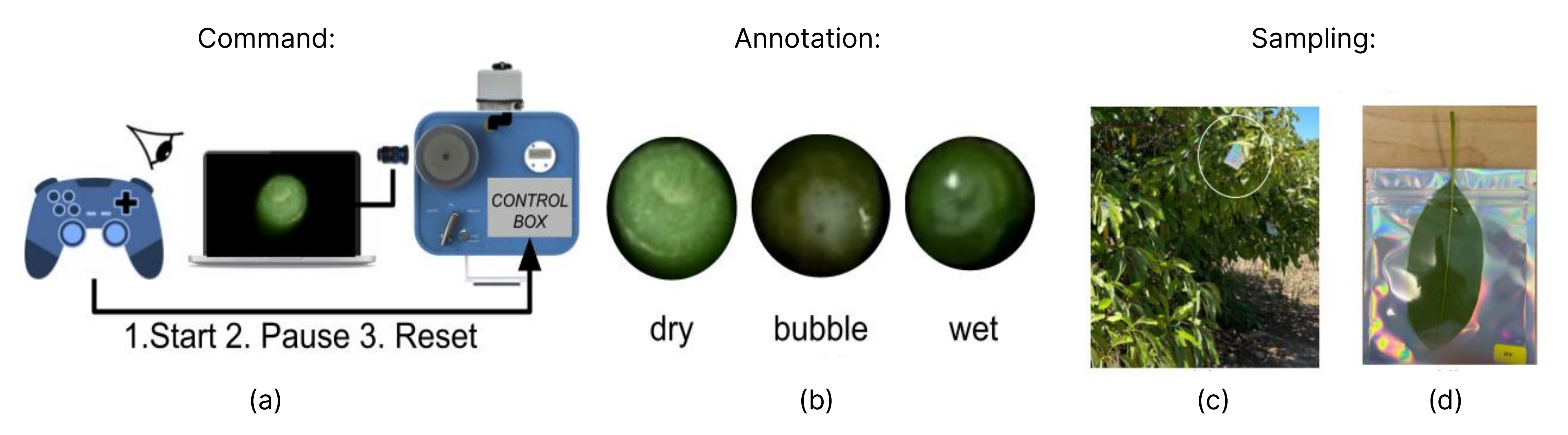}
    \vspace{-9pt}
    \caption{ (a) The data acquisition setup is based on an automated pressure chamber~\cite{mucchiani2024development}, and allows an operator full control of the process. (b) Examples of the three classes considered in this work. (c) Leaves have been bagged for 10 minutes before the excision from the tree. (d) The bagged leaf. 
    }
    \label{fig:framework2}
    \vspace{-12pt}
\end{figure}

We used the automated pressure chamber system developed by Mucchiani and Karydis~\cite{mucchiani2024development}. 
In summary, a commercially available pressure chamber (PMS 600D) is retrofitted with a single-board computer (Raspberry Pi Model 3B), a microcontroller (Arduino Uno R3) a camera (Raspberry Pi HQ camera V1), a joystick, an air compressor, a set of relays, a pressure sensor, and a solenoid valve (Fig.~\ref{fig:framework2}a). 
The system is rated to withstand a maximum pressure of $20\;bar$,  deemed sufficient for most avocado~\cite{bonomelli2019effect} and citrus~\cite{ortuno2006relationships} leaves, among others~\cite{chone2001stem}. 
Once a leaf is loaded (manually) by an operator, the process starts and the pressure inside the chamber increases, while simultaneously recording video data. 
The operator can pause, end, and reset the process at any point. 

\subsection{Data Collections}

We sampled leaves from local avocado trees located at UC Riverside's Agricultural Experimental Station (AES; $33\degree58'04.5''$N, $117\degree20'04.8''$W) following the procedure in~\cite{mucchiani2024development}. 
Healthy and mature leaves that receive full sun for part of the day were identified (Fig.~\ref{fig:framework2}c), placed in a retroreflective bag, left to settle for 10 minutes (Fig.~\ref{fig:framework2}d), and then excised cleanly at their stem. 
To promote sample variability and better test the methods' generalization capability, we conducted four field data collections and sampled leaves from multiple trees (with a maximum of five leaf samples from each tree in all but three field experiments) over different days and with different ambient temperature and soil moisture level (as deduced via the time elapsed post-irrigation). 
A total of $117$ leaves were collected; Table~\ref{tab:leaf} provides specific details. 

After sampling leaves, we performed SWP measurements. 
We manually focused the camera on the stem to get clear images (Fig.~\ref{fig:framework2}b). 
The aperture was adjusted to focus the lighting on the stem while minimizing reflections on other visible parts. 
The camera focus was occasionally re-adjusted to refine the view during experiments as needed. 
We ensured the pressure chamber was securely sealed before each experiment. 
Then, the pressure inside the chamber was increased until a clear xylem water expression. 
Once a clear ``wet'' xylem status was observed, the pressure was maintained constant for at least 2-3 seconds before releasing it.  
For safety purposes, the air compressor was stopped once the pressure reached $18$\;bar irrespective of the xylem wetness status. 
After complete pressure release, the recording was ended and the leaf was removed from the chamber. 
The camera was stationary throughout the entire process. 
Three researchers familiar with the process and hardware conducted these experiments.\footnote{In one instance we performed SWP measurements on two separate sessions. 
To keep the air compressor's temperature within safe limits, a maximum of 30 leaves can be tested at each SWP measurement session. 
Hence, the rest 20 leaves were stored in a cool and dry place overnight, and measurements resumed 23 hours post-excision. 
Images from the corresponding videos were used in stem detection; however, they were excluded from the xylem wetness classification as they were considerably drier.
}

\begingroup
\setlength{\tabcolsep}{3pt} 
\renewcommand{\arraystretch}{1.1} 
\begin{table}[!t]
\caption{Information on Leaf Data Collection and SWP Video Generation.}
\label{leaf_data}
\label{tab:leaf}
\centering
\begin{tabular}{c c c c c c}
\toprule
Date &  Time & Temp (F) & \# Trees & \# Leaves & Time elapsed post irrigation (h) \\
\midrule
05/01  & 11:00 am & 63 & 1 & 8  & unknown \\
05/05  & 12:00 pm & 61 & 10 & 50 & 55 (30 leaves) \& 78 (20 leaves)  \\
05/12  & 12:30 pm & 72 & 6 & 29 & 55.5 \\
07/03  & 11:00 am & 91 & 6 & 30 & 6 \\
\bottomrule
\end{tabular}
\vspace{-12pt}
\end{table}
\endgroup

\subsection{Dataset Generation and Curation}\label{method:dataset}
Videos from SWP measurements were processed to create three distinct datasets: the Stem Detection dataset, the Xylem Wetness Classification dataset, and the Video Evaluation dataset. 
The first two datasets comprise images extracted from SWP video footage cropped between several frames (but not overly many) before stem water expression and until the SWP cycle ends. 
Images were extracted at $30$\;fps with a resolution of $640\times480$ pixels and were stored in \textit{jpg} format. 
The \emph{Stem Detection dataset} comprises images from 30 videos of mostly dry measurements, collected on May 5th and July 3rd. 
The videos were split into training (18), validation (6), and testing (6) sets. 
Training and validation images were 4,500 and 1,500, respectively. 
For testing, we used the first 200 frames of the testing set videos. 
The \emph{Xylem Wetness Classification dataset} comprises images sampled from 40 videos collected on May 1st, 5th, and 12th. 
The videos were split into training (24), validation (8), and testing (8) sets. 
Frames were sampled so that a balanced distribution between all the classes was maintained, thus resulting in 13,500/4,500/4,500 images, respectively. 
Images in this dataset were cropped around the stem using bounding boxes derived via manual annotation and automated stem detection developed based on the Stem Detection dataset.

To determine the ground truth for classification, we initially annotated the xylem wetness into ``dry,'' ``wet,'' and ``bubble'' of the first 30 videos collected on May 5th. 
Every video was annotated independently by three different annotators. 
After some initial annotations and observed disagreements, the team discussed in depth to define ``wet'' as being more transparent, while ``bubble'' appears to be whiter, due to the scattering of light in different directions. \footnote{We refer the reader to \url{https://bit.ly/arcs\_isvc24} for supplementary information.}
The refined definition was used to guide the remaining annotations. 

The \emph{Video Evaluation dataset} comprises 20 videos collected on July 3rd that were used for end-to-end evaluation of the overall pipeline (i.e. joint stem detection and xylem wetness classification). 
Videos were annotated with both bounding boxes and wetness classification as in the other two datasets.

\vspace{-0.3cm}
\subsection{Stem Detection}
\label{sec:stem_det}
\vspace{-0.1cm}
We explored both classical vision-based methods and learning-based methods for stem detection. 
The distinct strengths and weaknesses of methods belonging to these two categories have been studied across different application contexts~\cite{o2020deep}. 
However, evaluating the efficacy of these tools within the context of this work (stem detection in the SWP process) is new, and it is thus important to understand how tools in both categories compare against each other.
Different methods were evaluated in terms of (1) the intersection over union (IoU) with respect to manually annotated ground truth bounding boxes, and (2) their execution time to successfully detect the bounding box around the stem.

Hough Transform~\cite{duda1972use} is a classical method used for line and curve detection. 
It has been used in agronomy to detect the diameter of tree trunks~\cite{li2023use}; that result inspired us to investigate it for stem detection. 
The OpenCV~\footnote{https://opencv.org/} built-in Hough Circle Transform function was used for detecting Hough circles across all frames. 
Any frames with multiple detections were discarded considering that exactly one stem is present in all cases. 
For the remaining frames, we rounded the center and radius of each circle to the nearest pixel and created a histogram of the position and radius. 
We then took the mode of all detections and evaluated the cases with the radius padded by 20 (H20), 30 (H30), and 40 (H40) pixels.

YOLOv5nu~\footnote{https://doi.org/10.5281/zenodo.3908559} and YOLOv8n~\footnote{https://github.com/ultralytics/ultralytics} are the two learning-based methods employed in this work. 
Their selection was based on their widespread use across applications. 
To create the YOLO-derived bounding boxes we used the coordinates corresponding to the top-left and bottom-right corners.

\subsection{Xylem Wetness Classification}
Given that different networks with good performance have been developed for classification tasks, we elected to follow a tiered approach. 
Specifically, from a large set of networks, we downselected the best-performing ones which were ultimately evaluated using video footage of SWP measurements. 

In the first phase, we considered fifteen CNN model architectures known to be computationally efficient and used in edge computing. 
These include EfficientNet B0/B1~\cite{tan2019efficientnet}, Inception V3~\cite{szegedy2016rethinking}, MobileNetv2~\cite{sandler2018mobilenetv2}, MobileNetv3-small~\cite{howard2019searching}, ResNet18/34/50~\cite{he2016deep}, Shufflenet v2 1.0/2.0~\cite{ma2018shufflenet}, SqueezeNet 1.0/1.1~\cite{iandola2016squeezenet}, VGG16/ 19 \cite{simonyan2014very}, and YOLOv8n-cls. 
All models were implemented using PyTorch~\cite{paszke2019pytorch} and were pretrained on the ImageNet~\cite{deng2009imagenet} dataset. 
We trained all models except YOLOv8n-cls for 20 epochs with a batch size of 16, a learning rate of $1e-3$, and via Stochastic Gradient Descent (SGD) optimizer. The images were resized as needed to meet each model's input image size requirement. 
YOLOv8n-cls was trained for 30 epochs with a $64\times64$ image size, 16 batch size, and 0.01 initial and final learning rate with SGD optimizer. 
We compared the Top-1 testing accuracy of the best and last trained model, the time to train an epoch, and whether the validation accuracy increased during the training process. 
The criteria to downselect networks to consider in the next phase included: (1) Top-1 accuracy greater than $70\%$, (2) training time less than $100$\;s, and (3) the epoch with the best validation accuracy being greater than one. 
In the case that multiple network variants belonging to the same family (e.g., the three ResNet variants) passed the bar, we kept only the best-performing one.

Seven networks moved on to the second phase. 
We then performed a hyperparameter search for those models (Table~\ref{tab:hyperparameter}).
We employed most of the built-in parameters for the downselected YOLOv8-cls network, tuned the optimizer (Optim), and varied the number of epochs (E) and the final learning rate (LR$_{final}$). 
These led to five different variants for YOLOv8-cls to be evaluated. 
For the remaining CNN models, we tuned the optimizer, learning rate (LR), learning rate scheduler (LRS) milestones for different number of epochs, and batch size (B). 
We also augmented (Aug) the dataset using random cropping and froze top layers during training to further improve model performance. 
These led to 18 different variants for each of the six other CNN networks to be evaluated. 
The models' best and average Top-1 testing accuracies were the evaluation metrics. 

\begingroup
\setlength{\tabcolsep}{3.5pt} 
\renewcommand{\arraystretch}{1.01}
\begin{table}[!t]
\centering
\caption{Hyperparameters Used for Tuning Downselected Models in Phase 2.}
\label{tab:hyperparameter}
\begin{tabular}{c c c c c c c c}
\toprule
& \multirow{2}{*}{Optim} & \multirow{2}{*}{LR} & LRS & Batch & \multirow{2}{*}{Epoch} & \multirow{2}{*}{Aug} & \multirow{2}{*}{LR$_{final}$}\\
 & & & Milestones & Size & & \\
\midrule
\multirow{3}{*}{CNN} & \multirow{1.75}{*}{SGD} & \multirow{3}{*}{1e-3} & None & 16  & 15  & \multirow{2.25}{*}{Random} & \\
 & \multirow{2.25}{*}{Adam} &   & [5, 10, 20-th epoch] & 32 & 20 & \multirow{1.75}{*}{Crop}& - \\
 &  &   & [1, 3, 5, 8-th epoch] & 64 & 30 &  & \\
 \midrule
\multirow{3}{*}{YOLOv8n-cls} & \multirow{1.75}{*}{SGD}  & \multirow{3}{*}{1e-2} &  &  & 10 & \multirow{3}{*}{Built-in} & \multirow{1.75}{*}{1e-2} \\
 &  \multirow{2.25}{*}{Adam} & & - & - & 30 &  &  \multirow{2.25}{*}{1e-6} \\
  &   & & &  & 50 &  &   \\
\bottomrule
\end{tabular}
\vspace{-12pt}
\end{table}
\endgroup

To further refine the evaluation, we chose the top two models in Top-1 accuracy and the top two in average Top-1 accuracy to move to the third phase. 
These four models were subsequently trained using two hyperparameter setups outlined in cases F1--F4 in Table~\ref{tab:ablation}. 
The best-performing configuration in terms of Top-1 accuracy for each of these four models was then moved on to the overall end-to-end framework evaluation with SWP video footage. 
Note that during the first two phases, we used cropped images based on the manually annotated stem bounding boxes. 
In the third phase, we used these, and in addition, the cropped images provided by the vision- and learning-based stem detection methods.

\begingroup
\setlength{\tabcolsep}{3pt} 
\renewcommand{\arraystretch}{1.01} 
\begin{table}[!t]
\centering
\caption{Hyperparameters Used in the Models Downselected for Phase 3.}
\label{tab:ablation}
\begin{tabular}{c c c c c c c c}
\toprule
Case/Criteria & B & E & LR$_{init}$ & LR$_{final}$ & LSR Milestones & Optim & Image Size\\
\midrule
F1 & 16 & 30 & 1e-3 & 1e-3 & None & SGD & $224\times224$ \\
F2 & 64 & 20 & 1e-3 & 1e-7 & [1, 3, 5, 8-th epoch] & SGD & $224\times224$ \\
F3 & 16 & 50 & 1e-2 & 1e-6 & - & SGD & $64\times64$ \\
F4 & 16 & 50 & 1e-2 & 1e-6 & - & SGD & $224\times224$ \\
\bottomrule
\end{tabular}
\vspace{-14pt}
\end{table}
\endgroup

Lastly, the final four downselected network configurations were combined with the downselected stem detection methods and were evaluated using the Video Evaluation dataset. 
During this evaluation, the stem detection method was run first on frames 501--700. 
Once the stem has been detected, the classification network determines xylem wetness from frame 701 onwards until all the frames are exhausted. 
The end-to-end framework's classification was evaluated using the Top-1 accuracy, Precision, Recall, and F1-score.


\section{Results and Discussion}
\subsection{Evaluation of Stem Detection} 
\begin{figure}[!ht]
    \centering
    \includegraphics[width=0.65\textwidth]{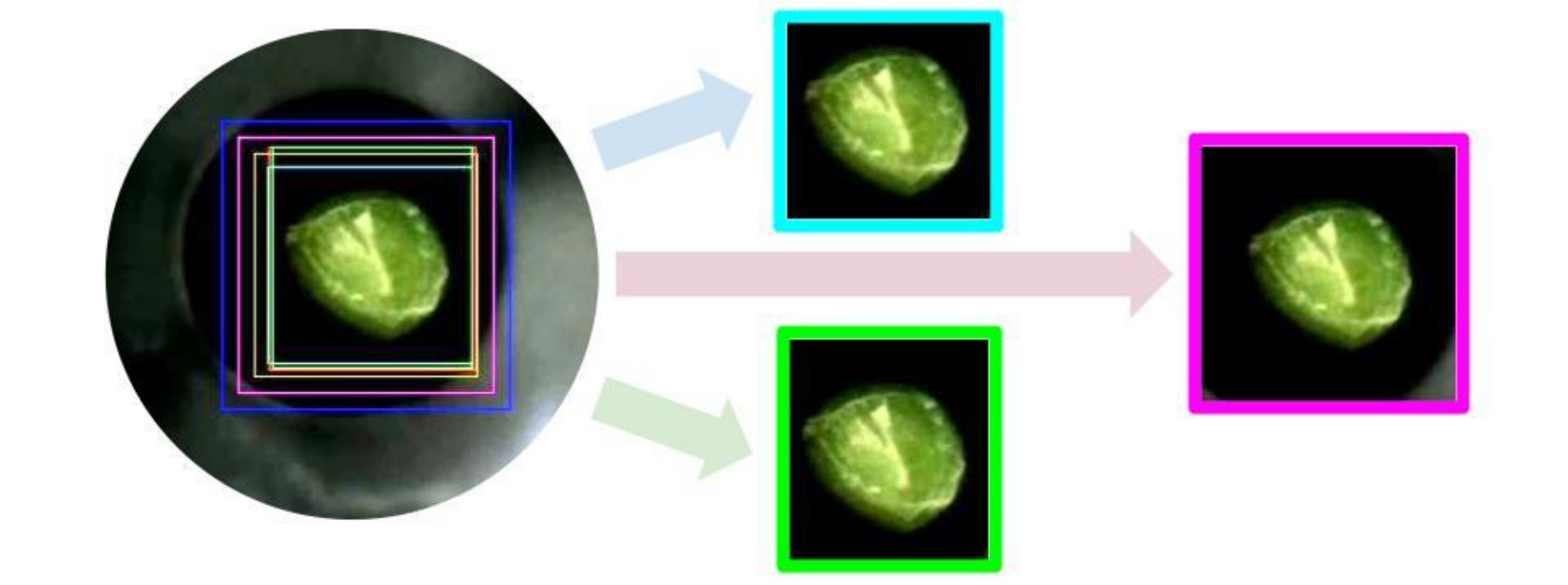}
    \vspace{-6pt}
    \caption{Sample stem detection bounding boxes via manual annotation (cyan), H20 (yellow), H30 (magenta), H40 (blue), YOLOv5nu (red), and YOLOv8n (green). The dimensions and placement of the bounding boxes differ between the tested methods. The two selected methods (H30 and YOLO8vn) along with the manually-annotated bounding box are highlighted.}
    \label{fig:stem}
    \vspace{-12pt}
\end{figure}

The Hough Transform with 20-pixel padding (H20) performs slightly worse but is generally comparable to YOLO network variants, with a maximum performance drop of about $5\%$ in terms of IoU (Table~\ref{tab:iou}). 
The execution time of the different Hough Transform paddings is nearly identical and approximately six times faster than that of YOLO, with both YOLO variants having similar execution times. 
The YOLO networks maintain high and consistent performance, with about a 3\% drop in IoU. 
The main difference is that YOLOv8n can provide tighter bounding boxes (see Fig.~\ref{fig:stem}). 
Therefore, YOLOv8n was chosen as the learning-based method to be tested in the overall end-to-end framework.

The Hough Transform can be quite fragile. 
As the padding increases the performance drops significantly (about $28\%$ max IoU drop between H20 and H40). 
This is likely because of the increased background noise added to the image part used in IoU computation, resulting in lower matching accuracy. 
Additionally, we tested the Hough Transform in the first 200 frames of the entire Stem Detection dataset (third row in Table~\ref{tab:iou}). 
In this case, there is higher variability in terms of the location of the stem in the provided images. 
This caused a significant drop in the H20 case (exceeding $10\%$), a small increase in H40, and a not significant change in the H30 case. 
These findings suggest that H30 is more stable, which is important for generalization, and was thus selected as the vision-based method to be tested in the overall end-to-end framework.

\begingroup
\setlength{\tabcolsep}{4pt} 
\renewcommand{\arraystretch}{1.1} 
\begin{table}
\vspace{-18pt}
\centering
\caption{Comparison of IoU for Different Stem Detection Methods.}
\label{tab:iou}
\begin{tabular}{c c c c c c}
\toprule
 &  H20 & {\bf H30} & H40 & YOLOv5nu & {\bf YOLOv8n}  \\
\midrule
Test IoU & 80.56     & 71.62 & 57.64  & 84.86  & 82.26 \\
time/video (s)     & 1.327    & 1.334 & 1.351  & 8.167  & 7.976\\
\midrule
\midrule
Full Stem Dataset IoU& 71.96     & 71.68 & 61.67  & -   & -\\
\bottomrule
\end{tabular}
\vspace{-24pt}
\end{table}
\endgroup

\subsection{Evaluation of Xylem Wetness Classification}\label{exp:classification}
Results from the evaluation of all 15 models considered initially are shown in Table~\ref{tab:phases}. 
\( T_E \) represents the time took to train the model for one epoch. 
\( P1_{BA} \) reports the Top-1 accuracy between the best and the last trained model. 
\( B_E \) is the epoch where the model demonstrated the best validation accuracy. 
A value of 1 indicates that the validation accuracy did not improve during training. 

Out of the 15 models, most exceeded the set threshold of $70\%$ Top-1 accuracy. 
Those models that failed to pass the bar (SqueezeNet 1.1 and VGG16/19) were able to classify only one class correctly (most often the ``bubble'' class). 
Several models with good Top-1 accuracy, however, did not show improvement past the first epoch of training and were thus discarded. 
Two exceptions were made to the latter case; Inception V3 and SqueezeNet 1.0. 
This is because both models had very high accuracy (over $90\%$ in classifying ``dry'' classes and marginal accuracy (over $65\%$) in classifying the other two classes; hence, we hypothesized that fine-tuning can further improve the performance of these models. 
Downselected models for the second phase testing are indicated with a checkmark in Table~\ref{tab:phases}.

\begingroup
\setlength{\tabcolsep}{5pt} 
\renewcommand{\arraystretch}{1.2}
\begin{table}[!ht]
\vspace{-18pt}
\centering
\caption{First Two Phases of  Xylem Wetness Classification Network Evaluation.}
\label{tab:phases}
\begin{tabular}{c c c c c c c c}
\toprule
\multirow{2}{*}{}CNN Model &  \( T_E \) & \( P1_{BA} \) & \( B_E \) & Phase 2 & \( P2_{BA} \) & \( P2_{AVG} \) & Phase 3 \\ & (s) \(\downarrow\) & \% \(\uparrow\) & & & \% \(\uparrow\) & \% & \\
 \midrule
EfficientNet B0   & 63.76     & 75.53 & 10  & \cmark  & 83.11  & \textbf{76.65}  & \cmark \\
EfficientNet B1    & 86.15     & 73.67 & 1  & \xmark   &  - & -  & \xmark\\
Inception V3        & 133.25    & 77.86 & 1  & \cmark  & 82.33  & 73.54  & \xmark\\
MobileNetv2     & 54.48     & 70.51 & 19& \xmark    &  - &  - & \xmark \\
MobileNetv3-small  & 45.47     & 89.07 & 2 & \cmark   &  \textbf{85.44} & 75.91  & \cmark\\
ResNet18        & 54.57     & 72.86 & 2 & \xmark   & -  & -  & \xmark\\
ResNet34         & 64.30     & 74.78 & 1 & \xmark    &  - & -  & \xmark\\
ResNet50           & 90.87     & 73.15 & 20& \cmark   &  \textbf{84.86} &  76.39 & \cmark\\
ShuffleNet v2 1.0   & 49.60     & 81.93 & 8 & \cmark   &  84.09 & 75.89  & \xmark\\
ShuffleNet v2 2.0  & 52.28     & 74.73 & 1 & \xmark    & -  & -  & \xmark\\
SqueezeNet 1.0     & 46.49     & 80.67 & 1 & \cmark   &  83.16 & 70.47  & \xmark\\
SqueezeNet 1.1      & 40.66     & 33.33 & 1 & \xmark    & -  &  - & \xmark\\
VGG16              & 136.07    & 33.33 & 1 &  \xmark   & -  & -  & \xmark\\
VGG19              & 150.05    & 33.33 & 1 &  \xmark   & -  &  - & \xmark\\
YOLOv8n-cls        &    43.12       & 82.42 & - &  \cmark  &83.15  & \textbf{78.79}  & \cmark\\
\bottomrule
\end{tabular}
\vspace{-12pt}
\end{table}
\endgroup

The different parameterizations (Table~\ref{tab:hyperparameter}) were then implemented for all downselected models. 
The overall best \( P2_{BA} \) and average \( P2_{AVG} \) Top-1 accuracy from these evaluations are listed in Table~\ref{tab:phases}. 
We noted that applying data augmentations did not improve training. 
Our datasets comprise images and videos that are tightly centered around the stem, and all look similar. 
As such, although there is some background, resizing and randomly cropping may lead to images that miss critical information. 
Further, the SGD optimizer led to better performance; this finding is consistent with other similar observations in the literature~\cite{zhou2020towards}. 
It was also observed that a faster decrease in the learning rate led to models with improved generalization accuracy. 
Overall, the difference in best performance between the seven fine-tuned networks is subtle (less than $4\%$ max difference in $P2_{BA}$). 
Differences in average performance are more pronounced (about $12\%$ max difference in $P2_{AVG}$).

\begingroup
\setlength{\tabcolsep}{3pt} 
\renewcommand{\arraystretch}{1.2} 
\begin{table}[!t]
\centering
\caption{Top-1 Accuracy of Models Integrated with Stem Detection Methods.}
\label{tab:final}
\begin{tabular}{c c c c c c c}
\toprule
CNN Model &  Manual crop & H30 crop & YOLO crop  \\
\midrule
EfficientNet B0 (F1)    & 81.29      &  76.2     &  77.44     \\
EfficientNet B0 (F2)    &  \textbf{81.64}     &  77.33     &  80.33     \\
\hline
MobileNetv3-small (F1) &  82.11     &   80.64    &  76.27     \\
MobileNetv3-small (F2)  &  81.11     &   81.98    &  \textbf{82.22}     \\
\hline
ResNet50 (F1)            &  82.24     &  82.69     &   77.31    \\
ResNet50   (F2)         &  76.18     &  \textbf{84.24}     &   70.13    \\
\hline
YOLOv8n-cls (F3)        &   77.38 &   77.76    &  76.48     \\
YOLOv8n-cls (F4)        &   \textbf{82.93}    &   81.42    &    82.4   \\
\bottomrule
\end{tabular}
\vspace{-12pt}
\end{table}
\endgroup

The top two networks in each metric category were then selected for further evaluation using two parametric setups out of those listed in Table~\ref{tab:ablation}. 
Downselected detectors were also integrated. 
The specific selections, as well as results from this third phase of evaluation, are reported in Table~\ref{tab:final}. 
The best-performing combinations for each method are highlighted. 
It can be observed that networks trained using the H30 stem detector have higher accuracy with learning rate scheduling (F2). 
This is because H30 crops the image around the stem more tightly than manual and YOLO-based crops which, in turn, leads to less background noise in the image. 
As our approach follows a more generalized learning method to create a network that performs better on unseen data, smaller learning rates (F2) contribute to achieving a higher accuracy~\cite{wilson2001need}. 
Overall, EfficientNet B0 and MobileNetv3-small appear to be more stable when hyperparameters vary, while YOLOv8n-cls is the network listed affected by the employed stem detection means. 
ResNet50 has the widest variation in Top-1 accuracy among the different parameterizations considered herein.

\subsection{Video Evaluation}
\begin{figure}[!t]
    \centering
    \includegraphics[width=0.9\textwidth]{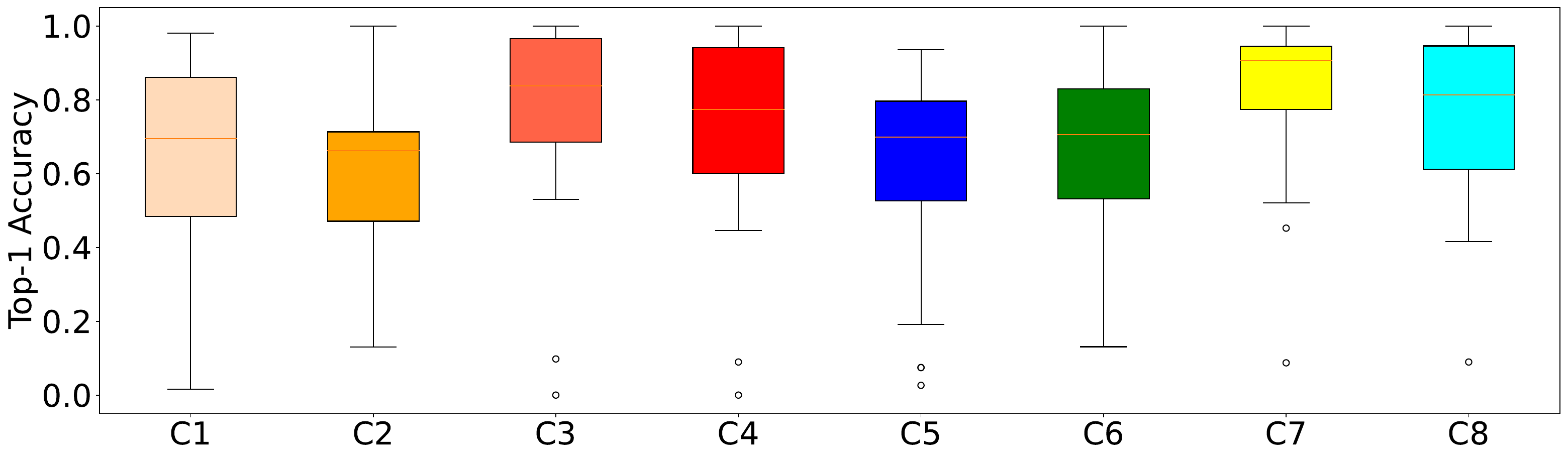}
    \vspace{-6pt}
    \caption{Top-1 Accuracy for end-to-end xylem wetness classification.}
    \label{fig:video_eval}
    \vspace{-12pt}
\end{figure}

The best-performing hyperparameter cases for each of the four networks in Table~\ref{tab:final} were combined with either the H30 or the YOLOv8n stem detector and were then evaluated in SWP video footage. 
Table \ref{tab:class} contains the results of this end-to-end xylem wetness classification pipeline over 20 different videos. 
It can be readily observed that YOLOv8n consistently performs better as a stem detector in comparison to H30. 
The case of the YOLOv8n detector integrated with the ResNet-50 classifier (C7) demonstrated the best performance (about $22\%$ on average higher Top-1 Accuracy in comparison to the other cases). 
This case synergistically leverages the strong generalization capabilities of both YOLO and ResNet-50. 
Even though ResNet-50 is a CNN framework with considerable depth, it effectively retains information owing to its residual learning capacity. 
Finally, Fig.~\ref{fig:video_eval} presents the distribution of the Top-1 Accuracy of all eight cases listed in Table~\ref{tab:class}. 
It can be observed that case C7 has the least amount of variability and the highest average value. 
Case C3 (which also uses ResNet-50 but the H30 detector) is also performing well overall and could be considered a variable alternative, but its best output is less strong compared to when using the YOLO-based detector. 
The outliers in the boxplots correspond to no detection in some cases.
Further investigation into these cases revealed that the video brightness was too low and hindered detection. 
In all, these findings suggest that a combination of a learning-based stem detection method and a deep CNN wetness classifier, when trained meticulously, can provide accurate predictions, even for previously unseen data.

\begingroup
\setlength{\tabcolsep}{3pt} 
\renewcommand{\arraystretch}{1.2} 
\begin{table}[!t]
\centering
\caption{End-to-end SWP Video Footage Evaluation.}
\label{tab:class}
\begin{tabular}{c c c c c c c}
\toprule
Case & Stem Detection and Classification &  Top-1 Acc. & Precision & Recall & F1-score  \\
\midrule
C1 & H30 \& EfficientNet B0 (F2)     &  60.13 & 87.66  & 60.13 & 64.07 \\
C2 & H30 \& MobileNetv3-small (F2)        & 60.70 & 86.27  & 60.70 & 60.37 \\
C3 & H30 \& ResNet50 (F2) & 74.93 & 88.89  & 74.93 & 77.15  \\
C4 & H30 \& YOLOv8n-cls (F4) & 71.33 & 84.89  & 71.33 & 72.13   \\
C5 & YOLOv8n \& EfficientNet B0 (F2)     &  60.35 & \textbf{92.41}  & 60.35 & 65.20  \\
C6 & YOLOv8n \& MobileNetv3-small (F2)        & 67.33 & 89.68 & 67.33 &   67.30  \\
C7 & YOLOv8n \& ResNet50 (F2) & \textbf{80.98} & 90.17  & \textbf{80.98} & \textbf{82.58}  \\
C8 & YOLOv8n \& YOLOv8n-cls (F4) & 74.98 & 89.94  & 74.98  & 76.06  \\
\bottomrule
\end{tabular}
\vspace{-3pt}
\end{table}
\endgroup

\section{Conclusion}
We introduced a framework for automated vision-based classification of xylem wetness during stem water potential (SWP) experiments. 
Our approach combined two parts; stem detection and xylem wetness classification. 
We conducted an extensive tiered evaluation of a total of two different methods for stem detection and 15 different methods for xylem wetness classification as well as different parameterizations thereof via hyperparameter tuning. 
Evaluation considered both images and video footage provided by SWP measurements. 
Our findings revealed several viable combinations of detectors and classifiers and established key similarities and differences. 
In sum, the combination of a YOLO-based detector and a ResNet-50 classifier was identified as the best-performing configuration, owing to the synergies afforded by those models. 

To the best of our knowledge, this is the first comprehensive study to perform joint stem detection and xylem wetness classification in support of SWP measurement via the pressure chamber, a process that is critical for precision irrigation. 
This work can support further automation of the pressure chamber method to measure SWP. 
It can also yield crucial information for similar precision agriculture efforts that seek to leverage the recent advances in visual learning. 
Future work will seek to include different plant species (e.g., citrus) as well as investigate the use of other sensing modalities (e.g., near-infrared imaging). 

\begin{credits}
\subsubsection{\ackname} 
We gratefully acknowledge the support of NSF \# CMMI-2326309, USDA-NIFA \# 2021-67022-33453, and The University of California under grant UC-MRPI M21PR3417. Any opinions, findings, and conclusions or recommendations expressed in this material are those of the authors and do not necessarily reflect the views of the funding agencies.

\subsubsection{\discintname}
The authors have no competing interests to declare that are
relevant to the content of this article. 
\end{credits}

\bibliographystyle{splncs04}
\bibliography{ref}
\end{document}